\documentclass[a4paper,fleqn]{cas-sc}

\usepackage[numbers,sort]{natbib}

\usepackage{amsthm}
\usepackage{amssymb}
\usepackage{amsmath}
\usepackage{booktabs}
\usepackage{multirow}
\usepackage{makecell}
\usepackage{url}
\usepackage{subcaption}
\usepackage{todonotes}
\usepackage{microtype}

\newtheorem{definition}{Definition}

\DeclareMathOperator*{\auc}{AUC_{borji}}
\DeclareMathOperator*{\kl}{KL_{div}}
\DeclareMathOperator*{\minf}{SIM}

\definecolor{lightgray}{gray}{0.9}

\begin{document}
\let\WriteBookmarks\relax
\def\floatpagepagefraction{1}
\def\textpagefraction{.001}
\shorttitle{Synthetic Ground-Truth Framework for the XAI Evaluation}
\shortauthors{Miquel Miró Nicolau et~al.}

\title[mode=title]{A Synthetic Ground-Truth Framework for the Evaluation of Explainable AI Methods}

\tnotemark[1]

\tnotetext[1]{This work has been partially supported by the Italian Project Fondo Italiano per la Scienza FIS00001966 ``MIMOSA'', by the European Community Horizon~2020 programme under the funding schemes G.A. 101120763 ``TANGO'', by the European Innovation Council project “EMERGE” (Grant No. 101070918) and project PID2023-149079OB-I00 funded by MICIU/AEI/10.13039/501100011033 and by ERDF/EU.}


\author[1]{Miquel Miró-Nicolau}[orcid=0000-0002-4092-6583]
\fnmark[1]
\ead{miquel.miro@uib.cat}
\ead[URL]{https://www.uib.cat/}

\credit{Conceptualization, Methodology, Software, Validation, Formal analysis, Investigation, Resources, Data Curation, Writing – Original Draft, Writing – Review \& Editing, Visualization.}

\affiliation[1]{organization={Universitat de les Illes Balears},
                addressline={Cra. de Valldemossa, km 7.5.}, 
                city={Palma},
                postcode={07122}, 
                country={Spain}}


\cortext[cor1]{Corresponding author}

\author[2]{Francesco Spinnato}[orcid=0000-0002-3203-6716]
\fnmark[2]
\ead{francesco.spinnato@unipi.it}
\ead[URL]{https://di.unipi.it/en/}

\credit{Conceptualization, Methodology, Software, Validation, Formal analysis, Investigation, Resources, Data Curation, Writing – Original Draft, Writing – Review \& Editing, Visualization.}

\author[2,3]{Riccardo Guidotti}[orcid=0000-0002-2827-7613]
\fnmark[2]
\ead{riccardo.guidotti@unipi.it}
\ead[URL]{https://di.unipi.it/en/}

\credit{Conceptualization, Supervision, Writing – Original Draft, Writing – Review \& Editing.}

\affiliation[2]{organization={Uiniversity of Pisa},
                addressline={Largo B. Pontecrovo, 3}, 
                postcode={56127}, 
                city={Pisa},
                country={Italy}}


\affiliation[3]{organization={ISTI-CNR},
                addressline={Via G. Moruzzi, 1}, 
                city={Pisa},
                postcode={56127}, 
                country={Italy}
                }


  


\begin{abstract}
Evaluating explainable Artificial Intelligence (XAI) methods is a challenging task due to the lack of reliable evaluation procedures and, in particular, the absence of ground truth explanations. In the literature, existing evaluation approaches typically assess explanations by measuring their fidelity with respect to the predictions of a black-box model. However, such evaluation strategies only quantify the degree to which an explanation reproduces the model’s output, without ensuring that the explanation correctly reflects the underlying decision process. As a consequence, different explanations may achieve similar fidelity scores while providing inconsistent or misleading interpretations of the model behavior.
In this paper, we propose a framework for the evaluation of XAI methods based on synthetic ground truth. The proposed approach relies on controlled interventions to generate synthetic datasets in which the importance of input components can be determined by design. This enables the construction of ground truth explanations that are directly aligned with the behavior of the model under analysis. The framework is instantiated across three data domains, namely binary images, tabular data, and time series, allowing a comprehensive assessment of explanation methods in heterogeneous settings.
Experimental results obtained by evaluating nine widely used XAI methods show significant limitations in current techniques and highlight the importance of synthetic, intervention-based benchmarks for a reliable assessment of explanation quality.
\end{abstract}




\maketitle

\section{Introduction}
\label{sec:intro}
EXplainable Artificial Intelligence (XAI) has emerged as a broad and rapidly evolving field, with applications spanning numerous domains~\cite{loh2022application,miro-nicolau2022evaluating}, including healthcare, finance, transportation, and public policy, where understanding the reasoning behind automated decisions is crucial for ensuring transparency, trust, and accountability~\cite{bodria2023benchmarking}. 
Despite the remarkable success of modern machine learning models, particularly deep neural networks, these systems are often regarded as \emph{black boxes}, as their internal decision-making processes remain largely opaque to human interpretation~\cite{guidotti2019survey}.
In response, a wide range of XAI methods has been proposed to provide insights into model behavior. 
However, a fundamental open challenge lies in the lack of agreement among different explanation techniques. 
This phenomenon, commonly referred to as the \emph{disagreement problem}~\cite{krishna2024disagreement}, raises critical concerns about the reliability, consistency, and ultimately the trustworthiness of explanations produced by XAI methods.

Despite this growing body of work, there is still a lack of standardized evaluation measures that enable a systematic and reliable assessment of explanation quality. 
A common strategy to address this limitation is the systematic evaluation of XAI methods, with the goal of identifying which explanations more faithfully reflect the behavior of the underlying model. 
Several authors~\cite{doshi-velez2018considerations,vilone2021notionsa,bodria2023benchmarking,nauta2023anecdotal} have proposed taxonomies to categorize XAI evaluation techniques. 
Broadly, these approaches can be divided into two main classes: \emph{human-grounded} and \emph{functionally-grounded} evaluations.
Human-grounded approaches rely on user studies to assess explanations, focusing on how humans interpret, understand, or interact with them. 
In contrast, functionally-grounded methods evaluate explanation quality through quantitative metrics without human involvement, and are therefore often referred to as machine-centered evaluations.

In this work, we focus on functionally-grounded evaluation techniques. 
Recent surveys~\cite{nauta2023anecdotal,dembinsky2026unifying} highlight both the rapid growth and the increasing fragmentation of this research area. 
In particular, Nauta \emph{et al.}~\cite{nauta2023anecdotal} identify twelve explanation qualities, six of which can be assessed using functionally-grounded approaches. Building on this perspective, Dembinsky \emph{et al.}~\cite{dembinsky2026unifying} conduct a large-scale systematic review, identifying more than 400 evaluation metrics, which are organized into 41 metric families and five major categories. A central conclusion of their analysis is the key role of \emph{fidelity} (also referred to as faithfulness), described as the ``foundation of the entire evaluation process''. At the same time, the large number and diversity of metrics reflect a lack of consensus on how explanation quality should be assessed.
Given this wide variety of evaluation approaches, several studies~\cite{tomsett2020sanity,hedstrom2023metaevaluation,miro-nicolau2025comprehensive,MIRONICOLAU2026132651} have undertaken \emph{meta-evaluations} of existing metrics, investigating whether they effectively measure the properties they are intended to capture. 
Although these works consider different subsets of metrics, they reach a consistent conclusion: many widely used evaluation measures exhibit limited reliability and unstable behavior across experimental settings. 
This limitation largely stems from the absence of an \textit{external} reference, such as a ground truth, forcing evaluation to rely on internal criteria, similarly to what occurs in the unsupervised evaluation of clustering algorithms~\cite{DBLP:books/aw/TanSK2005}. 
As a result, these approaches attempt to assess explanation quality based solely on intrinsic properties, without a direct notion of correctness.
Consequently, metrics based exclusively on \textit{internal} properties of explanations often provide only partial and potentially misleading insights into their validity.

An alternative to meta-evaluation approaches is the use of \emph{a priori} constraints that define a known ground truth for explanations. 
According to the taxonomy proposed by Dembinsky \emph{et al.}~\cite{dembinsky2026unifying}, these approaches can be broadly divided into two categories: human-annotated and synthetic datasets. 
In the latter case, controlled experimental settings are constructed so that the importance of each input feature is known by design. 
This enables the generation of ground truth (GT) explanations that can be directly compared with the outputs of XAI methods.
This class of approaches has recently been referred to as \emph{Synthetic Artificial Intelligence Ground Truth} (SAIG)~\cite{miro-nicolau2026exploring}. 
Within the SAIG paradigm, datasets are artificially generated to allow systematic manipulation of the contribution of specific input components. 
As a result, these methods provide reliable ground truth explanations without requiring human annotations, overcoming a key limitation of existing evaluation strategies.
Among these approaches, the framework proposed by Hesse \emph{et al.}~\cite{hesse2023funnybirds} has attracted attention due to its \textit{intervention-based design}, which enables the construction of ground truth explanations for the quantitative evaluation of XAI methods. However, the original formulation is restricted to color image data.

Building on these considerations, in this work we start from such intervention-based SAIG framework, and we extend it beyond the image domain. 
In particular, we investigate how the principles of intervention-based ground truth generation can be generalized to other data modalities commonly used in machine learning. 
To this end, we propose a unified methodology for constructing SAIG frameworks across binary images, tabular data, and time series, enabling the systematic and controlled evaluation of XAI methods in heterogeneous settings.
The main contributions of this work are threefold. 
First, we define the intervention-based SAIG framework for multiple data modalities, including binary images, tabular data, and time series. 
Second, we introduce, to the best of our knowledge, the first faithful \emph{local} ground truth construction for the evaluation of XAI methods in tabular data, addressing the limitations of existing approaches based on global feature importance. 
Third, we show how the SAIG framework can be tailored to time series data, enabling the evaluation of XAI methods in a domain where reliable ground truth has so far been largely unavailable.

The remainder of this paper is organized as follows. 
Section~\ref{sec:related} reviews the relevant literature on XAI evaluation. 
Section~\ref{sec:background} introduces the key concepts underlying the proposed framework. 
Section~\ref{sec:method} presents the methodology for constructing synthetic ground truth across binary images, tabular data, and time series. 
Section~\ref{sec:experimental} describes the experimental setup adopted to evaluate nine XAI methods across these data modalities. 
Section~\ref{sec:results} reports and discusses the experimental results. 
Finally, Section~\ref{sec:conclusions} concludes the paper and outlines directions for future research.

\section{Related Work}
\label{sec:related}
We review here existing approaches for the evaluation of XAI methods based on synthetic ground truth. 
These approaches construct controlled settings in which ground truth explanations are available by design, enabling a direct assessment of explanation quality. 
We focus on methods developed for tabular and image data, as, to the best of our knowledge, no SAIG frameworks for time series data have been proposed in the literature.

Several approaches have been proposed to construct synthetic ground truth for \textit{tabular data}. 
In~\cite{cortez2013using} is introduced one of the first SAIG methodologies by generating simple synthetic datasets with a limited number of input features. 
It relies on predefined functions $a: \mathbb{R}^M \rightarrow \mathbb{R}$ that assign weights to input variables, which are interpreted as feature importance. 
However, this formulation provides a \textit{global} notion of importance rather than instance-specific local explanations. 
Similar limitations are shared by subsequent methods that define ground truth through synthetic data generation processes~\cite{barr2020ground,liu2021synthetic,agarwal2022openxai}, where feature relevance is determined globally by construction.
Other approaches aim to evaluate specific explanation techniques. 
For instance, in~\cite{amiri2020data} proposed a SAIG framework tailored to LIME~\cite{ribeiro2016why}. 
While this method enables targeted evaluation, it lacks generality, as it is tightly coupled to a single explanation technique and relies on the assumption that LIME produces correct explanations, a claim that has been widely debated in the literature~\cite{lin2019explanations,ghorbani2019interpretation,slack2020fooling}.
A different perspective is in~\cite{guidotti2021evaluating}, who defines synthetic classifiers using rule-based systems instead of trained machine learning models. 
This setting allows full control over the decision process and provides a known ground truth for feature importance. 
However, such approaches are inherently limited to model-agnostic explanation methods that infer importance solely from input-output relationships, without access to the internal behavior of learned models.
Overall, existing methods for tabular data exhibit two main limitations: either \emph{(i)} they define ground truth independently of the model, potentially reducing adherence to its actual behavior~\cite{cortez2013using,barr2020ground,liu2021synthetic, agarwal2022openxai}, or \emph{(ii)} they are restricted to specific explanation techniques or settings, limiting their general applicability~\cite{amiri2020data,guidotti2021evaluating}.

Recent work has explored synthetic ground truth generation for \textit{image data}. In~\cite{miro-nicolau2026exploring} is provided a comprehensive overview of this area, identifying several representative approaches~\cite{yang2019benchmarking,arias-duart2022focus,hesse2023funnybirds,miro-nicolau2024assessing}.
In~\cite{yang2019benchmarking} is constructed composite images by combining foreground objects from MSCOCO~\cite{lin2014microsoft} with backgrounds from MiniPlaces~\cite{zhou2017places}, associating each image with dual labels corresponding to object and scene classification. 
In this setting, faithful explanations are expected to highlight different regions depending on the prediction task. 
The authors of~\cite{arias-duart2022focus} propose a mosaic-based approach, where each image is formed by combining patches from multiple classes. 
They introduce the ``Focus!'' score to quantify whether explanations concentrate on regions relevant to the predicted class.
The paper~\cite{miro-nicolau2024assessing} extends earlier tabular-based approaches to images by defining synthetic functions over image-derived features, such as region area or pattern frequency. 
However, these methods present several limitations, including a reliance on global rather than local importance~\cite{miro-nicolau2024assessing}, an emphasis on spatial localization rather than attribution magnitude~\cite{arias-duart2022focus}, and the use of comparative evaluations across models instead of intrinsic performance measures~\cite{yang2019benchmarking}.
A notable advancement is the intervention-based framework introduced in~\cite{hesse2023funnybirds}, which is evaluated on the \textit{FunnyBirds} dataset. This dataset consists of synthetic bird images composed of semantic parts, such as beaks and wings, and the class labels are determined by specific components.
By systematically intervening on these parts and observing the resulting changes in model predictions, the framework derives ground truth importance directly from model behavior. 
This intervention-based approach ensures a high degree of fidelity between the ground truth and the model’s decision process. 

\section{Preliminaries}
\label{sec:background}
In this section we introduce the notation and fundamental concepts used throughout this work. 
In particular, we formalize the machine learning model under analysis, the notion of local feature attribution explanations, and the intervention-based definition of feature importance used to construct ground truth explanations.
We begin by defining our input data. 
In particular, we consider three different data modalities, binary images, time series, and tabular data. 
For ease of notation each instance, independent of its data type, is represented as a vector $\mathbf{x} = [x_1,\dots,x_D]$, where $D$ denotes the total number of input features and $x_j$ represents the $j$-th scalar feature. 

\begin{definition}[Dataset]
A dataset, $\mathbf{X} = \{\mathbf{x}_1,\dots,\mathbf{x}_N\}$, is a collection of $N$ input samples, where each instance $\mathbf{x} \in \mathcal{X}$ belongs to an input space $\mathcal{X}$. 
\end{definition}

More specifically, for black and white images, each sample, $\mathbf{x}$, is represented as a $H \times W$ matrix, where $H$ and $W$ denote the image height and width, and $D=H\cdot W$ is the number of pixels. For time series, each instance is represented as a sequence $\mathbf{x} \in \mathbb{R}^{T}$, where $D=T$ denotes the number of timestamps. For tabular data, each sample corresponds to a feature vector $\mathbf{x} \in \mathbb{R}^{M}$, where $D=M$ denotes the number of tabular features. Based on the dataset definition above, we formalize the predictive model considered in this work, i.e., classification models.

\begin{definition}[Classification Model]
Let $\mathbf{X} = \{\mathbf{x}_1,\dots,\mathbf{x}_N\}$ be a dataset with instances $\mathbf{x} \in \mathcal{X}$ and labels $y \in \mathcal{Y}$. A classification model is a function $f : \mathcal{X} \rightarrow \mathcal{Y}$ that maps an input instance $\mathbf{x}$ to a predicted output $\hat{y} = f(\mathbf{x}).$
\end{definition}

In practice, $f(\mathbf{x})$ typically represents a vector of class scores or probabilities over the label space $\mathcal{Y}$, and the predicted class corresponds to the label with the highest score.
Independently of the data domain, state-of-the-art approaches for classification tend to be black-box models, not interpretable from a human standpoint~\cite{bodria2023benchmarking, theissler2022explainable,middlehurst2024bakeoff,spinnato2025pyrregular}.
For this reason, XAI approaches are increasingly used to shed light on the black-box decision-making.
Given a model $f$ and an input instance $\mathbf{x}$, the most common XAI methods explain individual predictions by assigning an importance score to each input feature via a so-called \emph{local feature attribution explanations}, as they describe the contribution of each feature to a single prediction~\cite{bodria2023benchmarking}. 

\begin{definition}[Local Feature Attribution Explanation]
A local feature attribution explanation for the prediction $f(\mathbf{x})$ is a vector $\mathbf{e} = [e_1,\dots,e_D] \in \mathbb{R}^{D}$ where each component $e_j$ represents the estimated importance of feature $x_j$ for the prediction associated with $\mathbf{x}$.
\end{definition}

The explanation vector $\mathbf{e}$ is produced by an attribution method $E$ applied to the model $f$ and the instance $\mathbf{x}$. 
In this work, we focus on \textit{post-hoc}, \textit{local}, \textit{model-agnostic} attribution methods, i.e., explanation techniques that operate externally to the black-box model and can therefore be applied to any predictive model regardless of its internal structure~\cite{bodria2023benchmarking}.
A common strategy for estimating feature importance is based on interventions on the input~\cite{ribeiro2016why,lundberg2017unified}. 
The basic idea is to measure how the model prediction changes when a particular input component is modified or removed. 
Importantly, these components do not necessarily correspond to individual scalar features, but may represent groups of features such as superpixels in images, temporal windows in time series, or subsets of variables in tabular data.
Let $\mathcal{S} = \{S_1,\dots,S_K\}$ denote a partition (or collection) of input components, where each $S_j \subseteq \{1,\dots,D\}$ is a subset of feature indices. 
Let $\mathbf{x} \setminus S_j$ denote an intervened version of the input where all features in $S_j$ have been altered according to a predefined intervention rule. 
The importance of $S_j$ can then be estimated as the change in the model output
$$e_j = f(\mathbf{x}) - f(\mathbf{x} \setminus S_j).$$

This formulation estimates the marginal contribution of the component $S_j$ to the prediction and is used in a wide range of \textit{post-hoc}, model-agnostic attribution methods. 
For instance, in image data, methods such as LIME~\cite{ribeiro2016why} and KernelSHAP~\cite{lundberg2017unified} define components $S_j$ as superpixels obtained through image segmentation. 
Perturbation-based methods such as RISE~\cite{petsiuk2018rise} implicitly define $S_j$ through random masks applied over the input. 
In time series, components $S_j$ typically correspond to contiguous temporal windows, as in methods such as T-SHAP~\cite{nguyen2025tshap}, or masked temporal segments as used in perturbation-based approaches. 
For tabular data, components $S_j$ are usually defined at the level of individual features or subsets of variables, as in LIME~\cite{ribeiro2016why} and KernelSHAP~\cite{lundberg2017unified}.

When evaluating explanation methods, it is important to evaluate how good are such local explanation, comparing them to a so-called ground truth. 
For better clarity, it is important to distinguish between two notions of ground truth. 
The first notion relates to the data-generating process, where certain features are assumed to be intrinsically responsible for the target label. 
The second notion, which is the focus of most XAI evaluation frameworks, concerns the behavior of the model itself. 
In this case, a ground truth explanation corresponds to the features that actually influence the model's prediction, regardless of whether the model has learned the correct underlying relationships.
In this work we focus on this second notion, under which the objective of an explanation method is to faithfully describe how the model produces its prediction, even if the model relies on spurious or undesirable correlations. 
We denote the corresponding ground truth attribution as $\mathbf{g} = (g_1,\dots,g_D)$, where each component $g_j$ represents the true contribution of feature $x_j$ to the prediction of the model.

Obtaining such ground truth explanations is challenging for real-world datasets. Performing interventions on natural data is often non-trivial, as removing a feature typically requires replacing it with an estimated value or perturbation, which may introduce artifacts or out-of-distribution samples. 
In this work, we exploit controlled synthetic settings in which such interventions can be performed without introducing artifacts, enabling the construction of reliable ground truth explanations aligned with the behavior of the model.

\begin{figure}[t]
    \centering
    \includegraphics[width=0.27\linewidth]{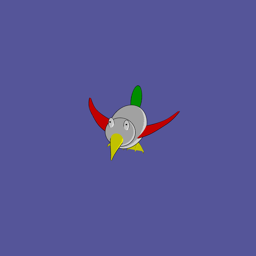}
    \includegraphics[width=0.27\linewidth]{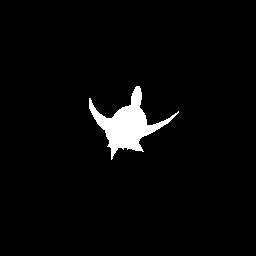}
    \includegraphics[width=0.35\linewidth]{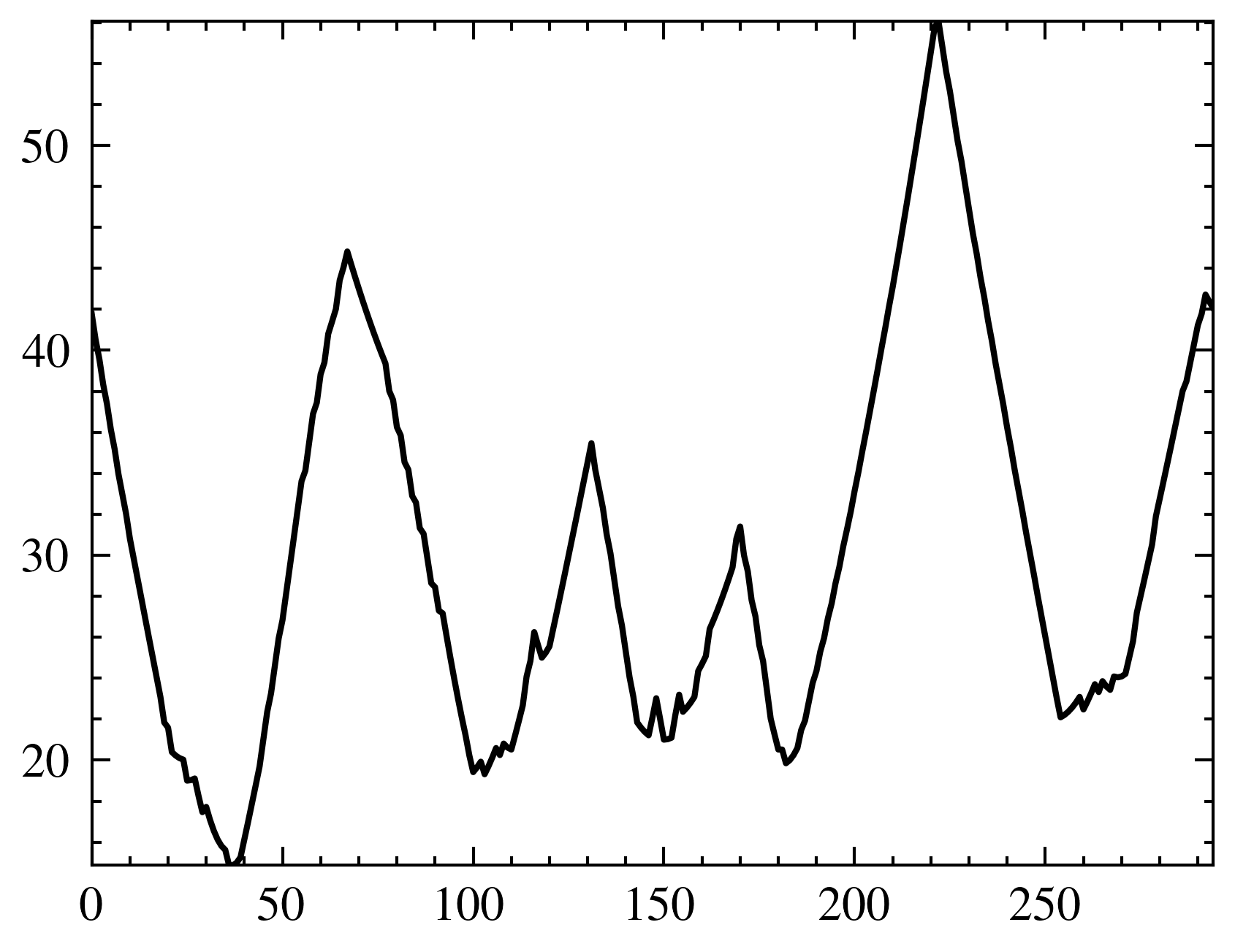}
    
    \includegraphics[width=0.27\linewidth]{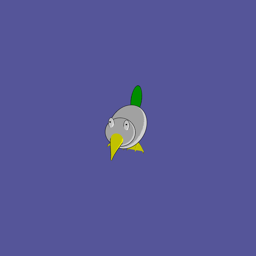}
    \includegraphics[width=0.27\linewidth]{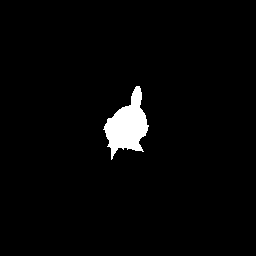}
    \includegraphics[width=0.35\linewidth]{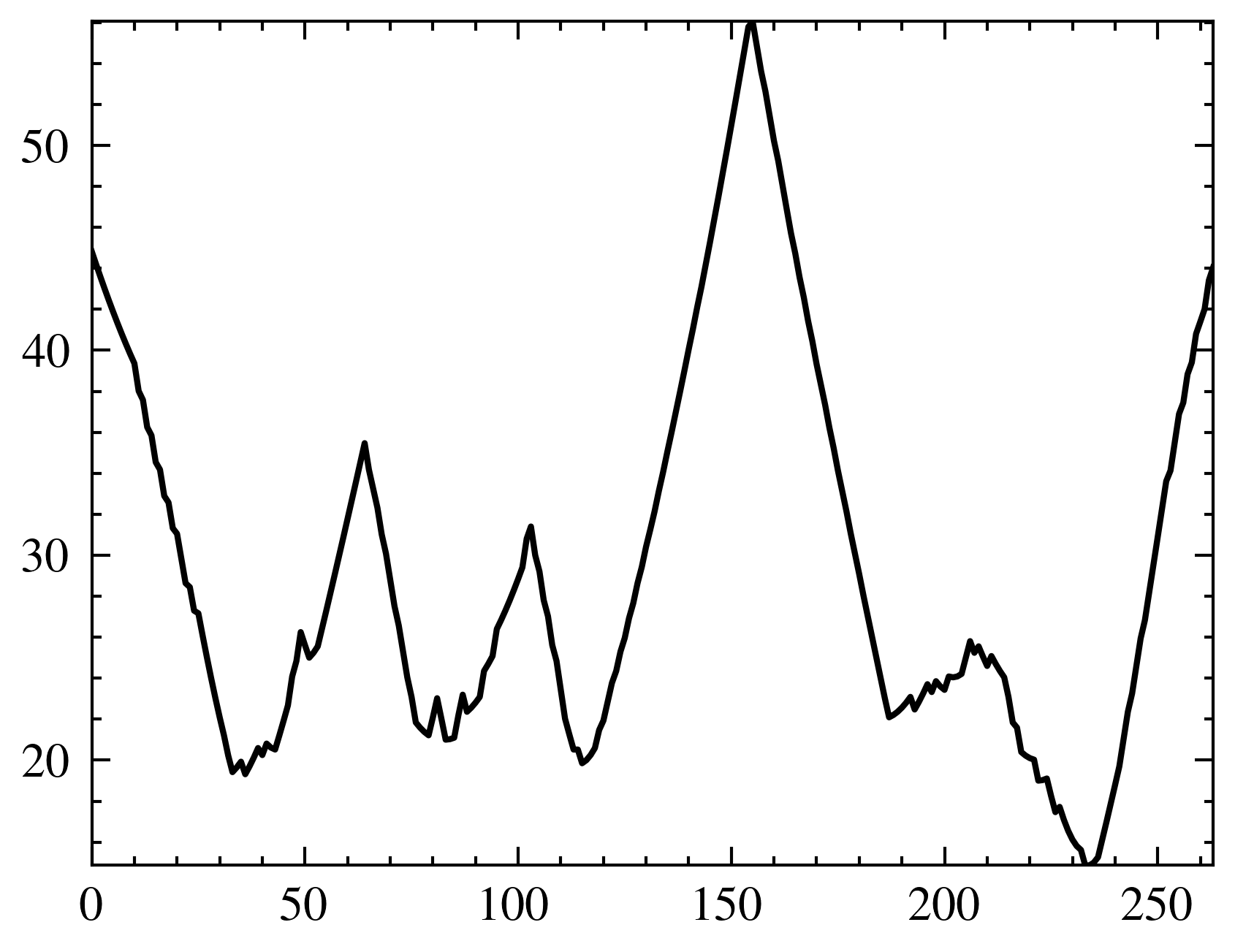}
    
    \caption{Illustration of the intervention procedure across data modalities: color images, binary images, and time series. The latter two are constructed from the original color image using the proposed framework.}
    \label{fig:intervent}
\end{figure}

\section{Methodology}
\label{sec:method}
In this section, we introduce the intervention-based approach for generating Synthetic Artificial Intelligence Ground Truth (SAIG) datasets across multiple data modalities, including binary images, tabular data, and time series. 
Figure~\ref{fig:intervent} illustrates the intervention procedure for each data typology: the first row shows the same original data with color, binary images and time series; the second row depicts the \textit{wings removal intervention} on all three typologies. 

First, we recall the intervention-based framework proposed for color images in~\cite{hesse2023funnybirds}.
Let represent an input image $\mathbf{x}$ as a collection of semantically meaningful components $\mathcal{S} = \{S_1, S_2, \dots, S_K\}$, where each component $S_j$ corresponds to a subset of pixels forming a distinct object part. 
These components are assumed to be disjoint, i.e., $S_i \cap S_j = \emptyset$ for $i \neq j$. 
The importance of each component $S_j$ is then computed through an intervention-based formulation:
\begin{equation}
\label{eq:intervention}
    g_j = f(\mathbf{x}) - f(\mathbf{x} \setminus S_j),
\end{equation}
where $g_j$ denotes the importance of component $S_j$, $f$ is the predictive model, and $\mathbf{x} \setminus S_j$ represents the intervened input obtained by removing the component $S_j$. 
In this setting, the intervention consists of replacing the pixels belonging to $S_j$ with the original background that the component occludes. Consequently, all pixels within a given component are assigned the same importance value.

\begin{figure}[t]
    \centering
    \includegraphics[width=0.30\linewidth]{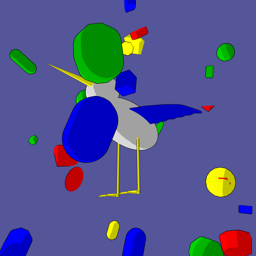}
    \includegraphics[width=0.30\linewidth]{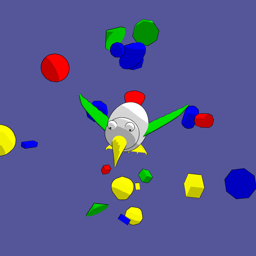}
    \includegraphics[width=0.30\linewidth]{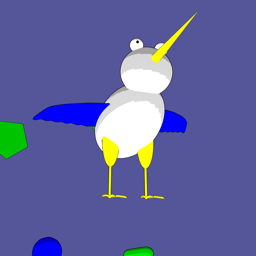}
    \caption{Representative samples from the FunnyBirds dataset across distinct classes. Each category is defined by a unique combination of geometric and color parts: wing color and foot shape (left image), wing and tail color (center image), and the foot shape (right image).}
    \label{fig:funny}
\end{figure}

In the specific case of the FunnyBirds dataset~\cite{hesse2023funnybirds}, each input $\mathbf{x}$ is composed of \textit{six} predefined semantic parts, namely \textit{beak}, \textit{eye}, \textit{feet}, \textit{tail}, \textit{wings}, and \textit{body}, each corresponding to a component $S_j \in \mathcal{S}$. 
The class label $y$ is determined by a specific combination of these parts.  M
ultiple variants are defined for each component, e.g., different shapes and colors. 
Figure~\ref{fig:funny} illustrates three real examples on how these variants combine, generating a large combinatorial space, from which a subset of 50 classes was selected.
The predictive model $f$ is trained to map each input $\mathbf{x}$ to its corresponding class label $y$. 
Specifically, the model is trained to predict classes directly from raw pixel data. Since no explicit semantic part information is provided, the model must autonomously learn to detect individual components and infer their relation to the predicted class.
A key aspect of this framework is that the training distribution is explicitly augmented with samples in which one or more components are removed. 
This design ensures that intervened inputs, such as $\mathbf{x} \setminus S_j$, remain within the training distribution, thereby mitigating potential out-of-distribution effects during evaluation.

In the following, we define the intervention-based SAIG framework to different binary images, tabular data, and time series. 
For each case, we define how to construct semantically meaningful components and how to perform interventions to preserve consistency with the underlying data generation process. 
This enables the computation of ground truth importance scores that remain faithful to the model behavior while ensuring that intervened samples lie within the training distribution. 
The proposed formulations maintain the core principles of the original framework while adapting them to the structural characteristics of each data modality.

\begin{figure}[t]
    \centering
    \includegraphics[width=0.24\linewidth]{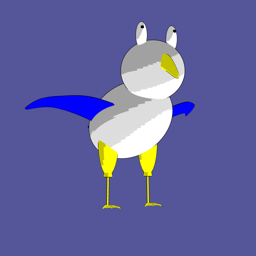}
    \includegraphics[width=0.24\linewidth]{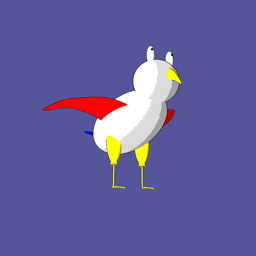}
    \includegraphics[width=0.24\linewidth]{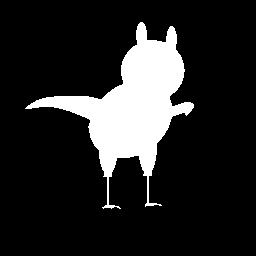}
    \includegraphics[width=0.24\linewidth]{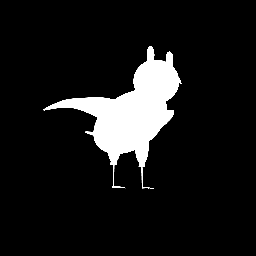}
    \caption{Illustration of class ambiguity induced by binarization. Two images belonging to different classes (first and second) become identical when converted to binary representations (third and fourth), as color information is removed.}
    \label{fig:similars}
\end{figure}

\subsection{Binary Images}
We define here the intervention-based SAIG framework for \textit{binary images}, which serves as a simplified domain for validating our methodology before extending it to more complex data modalities. 
In this setting, each input is represented as a binary mask $\mathbf{x} \in \{0,1\}^{H \times W}$, where foreground pixels (value $1$) indicate the presence of object components, and background pixels (value $0$) denote their absence. 
These binary representations can be interpreted as silhouette-like abstractions that preserve the geometric structure of the object while removing color information.

The construction of the binary dataset involves two main steps: binary mask generation and class redundancy pruning. First, we obtain binary images by aggregating the ground truth segmentation masks provided by the FunnyBirds framework, effectively mapping the original multi-channel representation into a single foreground-background mask. Formally, let $\mathbf{x}^{\text{rgb}}$ denote the original image and $\mathcal{S}=\{S_j\}_{j=1}^K$ the set of semantic components; the binary representation $\mathbf{x}$ is defined such that $x_{u,v} = 1$ if $(u,v) \in \bigcup_{j=1}^K S_j$, and $x_{u,v} = 0$ otherwise. This transformation preserves the spatial arrangement of components while eliminating color as a discriminative factor.

Second, we address the ambiguity introduced by the loss of color information. 
In the original dataset, several classes differ only in terms of color attributes and become indistinguishable after binarization. 
Figure~\ref{fig:similars} illustrates representative examples of such cases. 
To resolve this issue, we remove redundant classes while retaining a single representative for each group of shape-equivalent instances. 
Rather than merging classes, we adopt a pruning strategy to preserve a balanced class distribution. 
As a result, the total number of classes is reduced from 50 to 47.

The importance of each component is then computed using the same intervention-based formulation described previously. 
Given a component $S_j \in \mathcal{S}$, we construct the intervened input $\mathbf{x} \setminus S_j$ by replacing the corresponding pixels with the background value, i.e., setting them to zero. 
The contribution of $S_j$ is defined as the difference between the model outputs for the original and intervened inputs, consistently with the formulation in Eq.~(\ref{eq:intervention}).

\subsection{Time Series}
We extend the intervention-based SAIG framework to time series data by introducing a transformation that maps structured visual inputs into a one-dimensional sequential representation, while preserving the semantics of intervention-based importance. 
In~\cite{hesse2023funnybirds}, explanations are defined over spatial components, i.e., object parts, whose importance is quantified through controlled interventions. 
Our objective is to translate this paradigm into the temporal domain without losing the correspondence between components and their contributions.

\begin{figure}[t]
    \centering
    \includegraphics[width=1\linewidth]{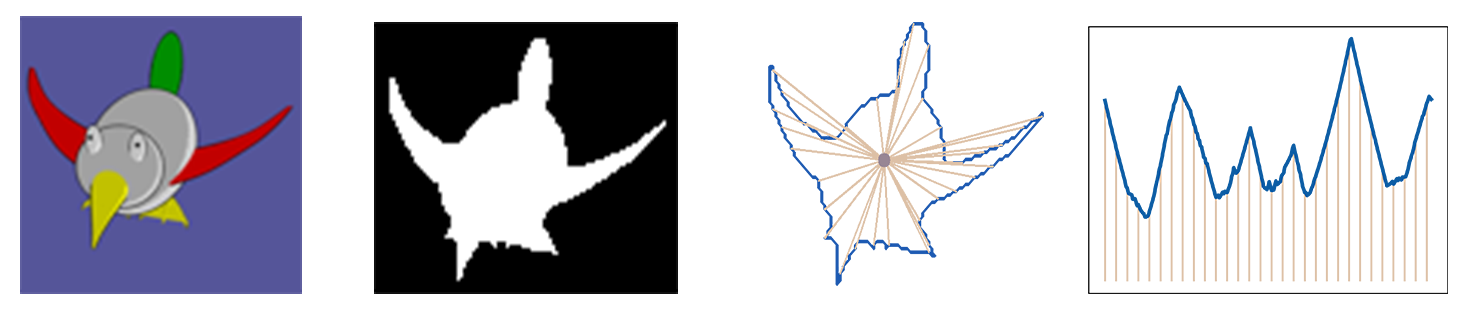}
    \caption{Step-by-step procedure for extracting a time series from a color image. From left to right: the original image, the binarized image, the measurement of radial distances from the object center to its contour, and the resulting time series plotted from these measurements}
    \label{fig:time_series}
\end{figure}

Figure \ref{fig:time_series} depicted the proposed two-step procedure to construct a time series from the original image.
First, we convert the original color images into binary representations, as described in the previous subsection, isolating the geometric structure of the object. 
Second, we transform the contour of the binary shape into a time series representation. Specifically, we extract the outer contour $\mathcal{C} = \{(u_t, v_t)\}_{t=1}^T$ of the object, where $(u_t, v_t)$ denotes the coordinates of the $t$-th contour point, and encode it as a univariate sequence by computing the Euclidean distance between each contour point and a reference centroid $\mathbf{c}$. The resulting time series $\mathbf{x} \in \mathbb{R}^T$ is defined as:
\begin{equation}
    x_t = \| (u_t, v_t) - \mathbf{c} \|_2, \quad t = 1, \dots, T,
\end{equation}
where $T$ denotes the length of the contour sequence. This representation is inspired by classical shape-to-sequence transformations in time series analysis~\cite{keogh2006lb_keogh}.
In our setting, the maximum sequence length $T$ is 1251 timestamps.
This transformation enables the reinterpretation of spatial components as contiguous temporal segments, allowing the application of the same intervention-based framework. 
Importantly, interventions are performed in the original input space by removing components from the image, and are subsequently propagated to the time series representation. 
Both the original input $\mathbf{x}$ and the intervened version $\mathbf{x} \setminus S_j$ are transformed using the same procedure, ensuring consistency between the resulting sequences.
To guarantee alignment across samples, we fix the centroid $\mathbf{c}$, computed as the mean of all object contour coordinates, from the original shape as the reference point for all corresponding interventions. 
This ensures that each temporal index $t$ corresponds to the same geometric location across interventions, preserving the coherence of the resulting importance estimates.
As a result, the proposed construction yields a time series SAIG in which the contribution of each temporal segment can be directly associated with the removal of a corresponding structural component in the original input.

\subsection{Tabular Data}
The evaluation of XAI methods for tabular data has been extensively studied in the literature, as discussed in Section~\ref{sec:related}. 
However, existing approaches exhibit two main limitations: they either provide a global notion of ground truth explanation, which is not suitable for assessing local methods~\cite{cortez2013using,barr2020ground, liu2021synthetic, agarwal2022openxai}, or they are not generalizable across different explanation techniques~\cite{amiri2020data,guidotti2021evaluating}. 
To address this gap, we propose an intervention-based SAIG framework for tabular data, combining it with the functional benchmark introduced in~\cite{cortez2013using}.

Cortez and Embrechts~\cite{cortez2013using} proposed a synthetic benchmark based on attribution functions, i.e., deterministic mappings $a : \mathbb{R}^{D} \rightarrow \mathbb{R}$ from which feature importance can be derived \emph{a priori}. 
A canonical example is a weighted linear function:
\begin{equation}
\label{eq:wlf}
    a(\mathbf{x}) = \sum_{j=1}^{D} w_j \, x_j,
\end{equation}
where $\mathbf{x} = [x_1,\dots,x_D]$ and $w_j$ denotes the weight associated with feature $x_j$. 
The output $a(\mathbf{x})$ is used as the target label for training a predictive model. 
In this formulation, feature importance is implicitly defined by the functional form of $a$. 
However, this assumption does not generally hold, particularly in non-linear settings where feature contributions may vary across instances.
To overcome this limitation, we introduce an intervention-based formulation. 
In order to perform controlled interventions while avoiding out-of-distribution effects, we define a discretized input space:
\begin{equation}
\label{eq:disc_input_space}
    \mathcal{X} = \left\{ \frac{k}{K} \;\middle|\; k \in \{0,1,\dots,K\} \right\}^{D},
\end{equation}
where $K$ denotes the number of discrete levels per feature and $D$ is the dimensionality of the input space.

Given an input $\mathbf{x} \in \mathcal{X}$, we define an intervention on feature $x_j$ as a decrease of one discrete level, yielding an intervened instance $\mathbf{x} \setminus_j$. 
The importance of feature $x_j$ is then computed as:
\begin{equation}
    g_j = f(\mathbf{x}) - f(\mathbf{x} \setminus_j).
\end{equation}
This formulation estimates the marginal contribution of feature $x_j$ to the model output.

The proposed intervention strategy provides several advantages. First, it avoids the need to specify an external baseline value. Second, due to the discretization of $\mathcal{X}$, the perturbed instance $\mathbf{x} \setminus_j$ is guaranteed to remain within the valid input domain, preventing out-of-distribution artifacts. 
Third, the importance values are directly derived from the model behavior, ensuring consistency between the explanation and the underlying predictive mechanism.
Fourth, this framework can be applied to all attribution functions introduced in~\cite{cortez2013using}, as well as to more complex functional benchmarks.

\section{Experimental Setup}
\label{sec:experimental}
In the previous section, we introduced a set of intervention-based SAIG frameworks to construct ground truth explanations for binary images, time series, and tabular data. 
Building on these formulations, this section describes the experimental protocol adopted to evaluate the fidelity of different XAI methods with respect to the proposed ground truth. 
We detail the evaluation metrics, the predictive models used in each domain, and the set of explanation methods considered in our study.

\subsection{Evaluation Metrics}
The proposed SAIG framework provides ground truth explanations that enable a direct comparison with the outputs of XAI methods. 
To quantify this comparison, it is necessary to adopt evaluation metrics that are consistent with the nature of feature attribution explanations. 
While this work focuses on feature importance, the framework can be extended to other explanation modalities by selecting appropriate evaluation measures.

A wide range of metrics has been proposed in the literature to compare feature importance maps with a reference ground truth, ranging from simple overlap-based measures such as Intersection over Union~\cite{oramas2019visual} to task-specific metrics~\cite{yang2019benchmarking,arias-duart2022focus,hesse2023funnybirds}. 
However, the problem of comparing attribution maps has been extensively studied in the context of visual saliency and human gaze prediction. 
In~\cite{riche2013saliency} is provided a comprehensive analysis of existing metrics and recommend a set of complementary measures based on their empirical properties.

Following these recommendations, we adopt three evaluation metrics.
For each evaluation metric $e_j$ and $g_j$, as defined in Section~\ref{sec:background}, denotes the estimated importance of the component $j$ its actual importance.

\paragraph{Similarity ($\minf$).} To quantify the agreement between ground truth and predicted importance distributions while preserving their spatial structure, we adopt the similarity metric ($\minf$)~\cite{judd2012benchmark}. 
It is defined as:
\begin{equation}
    \minf(e, g) = \sum^D_{j=1} \min (e_{j}, g_{j}).
    \label{eq:sim}
\end{equation}

\paragraph{Kullback--Leibler Divergence ($\kl$).} To measure the discrepancy between ground truth and predicted importance distributions in terms of information loss, we adopt the Kullback--Leibler divergence ($\kl$)~\cite{kullback1951information}. 
Unlike SIM, this metric does not explicitly account for spatial structure. 
It is defined as:
\begin{equation}
    \kl = \sum^{D}_{j=1} g_j \cdot \log \left( \frac{g_j}{e_j + \epsilon} + \epsilon\right).
    \label{eq:kl}
\end{equation}

\paragraph{Area Under the Curve ($\auc$).} 
To evaluate the ability of attribution methods to discriminate relevant from non-relevant regions while mitigating center-bias effects, we adopt the $\auc$ metric~\cite{borji2013quantitative}. This measure is a variant of the traditional ROC-AUC~\cite{green1966signal} in which negative samples are drawn from random locations rather than uniformly across the domain. 
It is defined as:
\begin{equation}
    \label{eq:aucb}
    \begin{aligned}
        \auc &= \int_{0}^{1} \text{TPR}(\tau) \cdot \text{FPR}(\tau), &
        \text{TPR}(\tau) &= \frac{|\{ j \in g \mid e_i \geq \tau \}|}{|g|}, &
        \text{FPR}(\tau) &= \frac{|\{ j \in \mathbf{R} \mid e_j \geq \tau \}|}{|\mathbf{R}|}.
    \end{aligned}
\end{equation}
where $\mathbf{R}$ is a set of randomly sampled elements of the vectors.
We employ $\minf$ and $\kl$ across all experimental domains. 
The $\auc$ metric is used only for image and time series data, as tabular data lack an inherent spatial structure, making such metrics not applicable in this setting.

\subsection{Predictive Models}
To assess the effectiveness of the proposed SAIG framework, we train a predictive model for each data modality using the datasets generated as described in Section~\ref{sec:method}. 
These models serve as test beds for evaluating whether the ground truth explanations derived from our framework enable a reliable assessment of XAI methods. 
While the design of predictive models is not the primary focus of this work, their role is essential to ensure that the evaluation reflects the behavior of learned models across different data domains.
We outline here the training procedures adopted in our study.

\paragraph{Binary Images Model.}
For the synthetically generated binary image datasets, we adopt a ResNet50 architecture~\cite{he2016deep} pre-trained on ImageNet~\cite{5206848}. 
To tailor the model to our task, we replace the original classification head with two fully connected layers and incorporate dropout regularization to mitigate overfitting. 
The first layer consists of 128 neurons, followed by a Rectified Linear Unit (ReLU) activation and a dropout layer with rate $p = 0.7$. 
The final output layer contains a single neuron with a Softmax activation function to produce class probabilities. 
During fine-tuning, the convolutional backbone is frozen, and only the newly introduced layers are updated.

The model is trained using the Adam optimizer with a learning rate of $10^{-5}$ and optimized with a cross-entropy loss function. 
Training is performed for 100 epochs with a batch size of 100. 
The training data consist of the binary image dataset generated according to the proposed methodology. 
In total, the dataset contains 47,470 images of size $256 \times 256$, of which 470 are reserved for validation (10 per class). 
The class distribution is consistent with that of the original dataset.

\paragraph{Time Series Model.}
For the synthetically generated time series datasets, we adopt a feature-based classification approach based on the MultiRocket transformation~\cite{tan2022multirocket}, a recent state-of-the-art method for efficient and accurate time series classification. 
MultiRocket extracts a large number of features by applying random convolutional kernels and pooling operations over temporal signals, resulting in a high-dimensional representation that captures diverse temporal patterns.
As a preprocessing step, sequences of variable length are aligned through padding. 
Missing values at the end of each sequence are filled by propagating the last observed value forward, ensuring that all time series have a consistent length and can be processed uniformly. 
The MultiRocket transformation is then applied to the preprocessed data to generate a large set of features, which are subsequently normalized to ensure comparable contributions during classification.

For the classification stage, we employ a linear ridge classifier. Prior work on the ROCKET family of methods has shown that ridge-based linear models achieve strong performance when applied to such high-dimensional feature representations~\cite{dempster2020rocket,tan2022multirocket}. 
Since ridge classifiers produce decision scores rather than probabilities, we apply a calibration step based on a sigmoid mapping, learned through cross-validation, to obtain calibrated probability estimates~\cite{guo2017calibration}. 
The regularization parameter is selected via cross-validation over a logarithmically spaced range between $10^{-3}$ and $10^{3}$.
The resulting pipeline consists of four stages: sequence alignment via padding, feature extraction using MultiRocket, feature normalization, and classification with a regularized linear model followed by probability calibration. 
This configuration provides an efficient and competitive solution for large-scale time series classification while maintaining low computational complexity.

\paragraph{Tabular Data Model.}
For the tabular dataset, we adopt a Multilayer Perceptron (MLP) architecture. 
The model consists of three fully connected layers, each with 128 neurons, followed by Rectified Linear Unit (ReLU) activation functions to introduce non-linearity. 
Given the binary classification setting, the final output layer uses a sigmoid activation function to produce class probabilities.
The network is trained using the Adam optimizer with an initial learning rate of $10^{-4}$. 
To improve convergence, we employ a step learning rate scheduler (StepLR), which reduces the learning rate by a factor of $\gamma = 0.85$ every 150 epochs. 
The model is optimized using a binary cross-entropy loss function over 5000 epochs with a full-batch training strategy.

The training dataset is generated according to the proposed SAIG methodology, using $K=100$ different discrete levels, and consists of approximately 4 million samples with three input features. 
Of these, 10,000 samples are reserved for test, while the remainder are used for training. 
The target labels are derived from the \textit{ssin-2c} attribution function introduced in~\cite{cortez2013using}:
\begin{equation}
    \textit{ssin-2c}(x) = \left[ w_1 \cdot \sin\left(\frac{\pi}{2} x_1\right) + w_2 \cdot \sin\left(\frac{\pi}{2} x_2\right) + w_3 \cdot \sin\left(\frac{\pi}{2} x_3\right) \right] > 0.6,
\end{equation}
where $w = [0.5, 0.25, 0.125]$. 
The output of $\textit{ssin-2c}(x)$ is used as the training label.

\begin{table}[t]
    \centering
    \begin{tabular}{ccccc}
    \toprule
    Data Modality       & Model & Precision & Recall        & F1-Score   \\      
    \midrule
    Binary images       & ResNet50       & $0.91$    & $0.88$        & $0.89$    \\
    Time series         & Multirocket w. Ridge      & $0.64$    & $0.63$        & $0.63$    \\
    Tabular data        & MLP      & $0.98$    & $0.98$        & $0.98$    \\
    \bottomrule
    \end{tabular}
    \caption{Performance measure results for each AI model used in each experiment.}
    \label{tab:perform}
\end{table}

\paragraph{Predictive Performance.}
The predictive performance obtained for each model is reported in Table~\ref{tab:perform}. 
We observe near-perfect performance in the tabular setting, while lower accuracy is obtained for binary images and time series data. 
However, this does not limit the validity of our evaluation, as the intervention-based procedure provides ground truth importance independently of model accuracy. 
Therefore, even for imperfect models, the expected contribution of each input component remains well-defined\footnote{The implementation details, including code and trained model weights, are publicly available at \url{https://github.com/miquelmn/fb2ts}.}.

\begin{table}[t]
\centering
\begin{tabular}{@{}lll@{}}
\toprule
\textbf{Method} & \textbf{Hyperparameter} & \textbf{Value} \\ \midrule


\multirow{3}{*}{RISE} 
    & N Masks                    & $600$ \\
    & S Masks                    & $8$ \\
    & Probability remaining      & $0.1$ \\ \addlinespace

\multirow{7}{*}{LIME}
    & Occlusion Value            & $0$ \\
    & N Samples                  & $1500$ \\
    & Kernel Width               & $0.25$ \\
    & Feature Selection          & Highest Weights \\
    & Kernel                     & Similarity kernel \\
    & Distance Metric            & Cosine \\
    & Model Regressor            & Ridge regression \\ \addlinespace

Gradient
    & Absolute value             & True \\ \addlinespace

\multirow{2}{*}{DeepLIFT}
    & Baseline                   & $0$ \\
    & $\epsilon$                 & $1 \cdot 10^{-10}$ \\ \addlinespace

\multirow{2}{*}{KernelSHAP}
    & Occlusion Value            & $0$ \\
    & N Samples                  & $25$ \\ \addlinespace

\multirow{3}{*}{Integrated Gradients}
    & N Steps                    & $50$ \\
    & Method                     & Gauss--Legendre quadrature \\
    & Baseline                   & $0$ \\ 

\bottomrule
\end{tabular}
\caption{XAI method hyperparameter values for binary image experiments.}
\label{tab:parameters_bin}
\end{table}

\begin{table}[t]
\centering
\begin{tabular}{@{}lll@{}}
\toprule
\textbf{Method} & \textbf{Hyperparameter} & \textbf{Value} \\ \midrule

\multirow{4}{*}{RISE}
    & N Masks                    & $1000$ \\
    & Grid resolution            & $16$ \\
    & Cell activation probability & $0.5$ \\
    & Mask generation            & Linear upsampling + random shift \\ \addlinespace

\multirow{3}{*}{LIME}
    & N Samples                  & $5000$ \\
    & Sampling                   & Gaussian \\
    & Discretize continuous features & False \\ \addlinespace

\multirow{2}{*}{KernelSHAP}
    & Background dataset         & Balanced centroid \\
    & Feature representation     & Time steps as features \\ \addlinespace

\multirow{4}{*}{T-SHAP}
    & Variants                   & ROI, Window \\
    & Window length              & $10\%$ of time series length \\
    & Stride                     & $5$ \\
    & Background dataset         & Balanced centroid \\ 

\bottomrule
\end{tabular}
\caption{XAI method hyperparameters values for time series experiments.}
\label{tab:parameters_ts}
\end{table}

\begin{table}[t]
\centering
\begin{tabular}{@{}lll@{}}
\toprule
\textbf{Method} & \textbf{Hyperparameter} & \textbf{Value} \\ \midrule
\multirow{7}{*}{LIME}
    & Occlusion Value            & $0$ \\
    & N Samples                  & $1500$ \\
    & Kernel Width               & $0.25$ \\
    & Feature Selection          & Highest Weights \\
    & Kernel                     & Similarity kernel \\
    & Distance Metric            & Cosine \\
    & Model Regressor            & Ridge regression \\ \addlinespace

Gradient
    & Absolute value             & True \\ \addlinespace

\multirow{2}{*}{DeepLIFT}
    & Baseline                   & $0$ \\
    & $\epsilon$                 & $10^{-10}$ \\ \addlinespace

\multirow{2}{*}{KernelSHAP}
    & Occlusion Value            & $0$ \\
    & N Samples                  & $25$ \\ \addlinespace

\multirow{3}{*}{Integrated Gradients}
    & N Steps                    & $50$ \\
    & Method                     & Gauss--Legendre quadrature \\
    & Baseline                   & $0$ \\ \addlinespace

\multirow{3}{*}{LORE}
    & Size                       & $1000$ \\
    & N Generation               & $10$ \\
    & Other Class Ratio          & $0.1$ \\

\bottomrule
\end{tabular}
\caption{XAI method hyperparameters values for tabular data experiments.}
\label{tab:parameters_tab}
\end{table}

\subsection{Explanation Methods}
Following the definition of the evaluation methodology, we describe the XAI techniques considered in our comparative analysis. 
Our study focuses on local feature attribution methods, which explain individual predictions by assigning an importance score to each input feature, rather than characterizing the model at a global level. 
In total, we evaluate nine approaches: RISE~\cite{petsiuk2018rise}, LIME~\cite{ribeiro2016why}, Gradient~\cite{simonyan2014deep}, DeepLIFT~\cite{shrikumar2017learning}, KernelSHAP~\cite{lundberg2017unified}, Integrated Gradients (IG)~\cite{sundararajan2017axiomatic}, T-SHAP~\cite{nguyen2025tshap} (in both ROI and Window variants), and LORE~\cite{guidotti2019factual}. 
Implementations for Gradient, DeepLIFT, KernelSHAP, and Integrated Gradients are based on the Captum library~\cite{kokhlikyan2020captum}, while for RISE~\cite{risegit}, LIME~\cite{limegit}, LORE~\cite{loregit}, and T-SHAP~\cite{nguyen2025tshap}, we use the official implementations provided by the respective authors.

For the binary image experiments, we evaluate RISE, LIME, Gradient, DeepLIFT, KernelSHAP, and Integrated Gradients. {\color{black} Table \ref{tab:parameters_bin} depicted each method hyperparameters used for binary images.}
RISE is configured with 600 random masks, a mask resolution of 8, and a retention probability of 0.1. 
For LIME and KernelSHAP, interpretable features are defined using the Quickshift algorithm~\cite{vedaldi2008quick} to generate superpixels. 
LIME is applied with 1500 perturbed samples, kernel width 0.25, cosine distance, ridge regression as surrogate model, and highest-weight feature selection. 
KernelSHAP uses an occlusion value of 0 and 25 samples. 
For gradient-based methods, Gradient is computed as the absolute value of the saliency map, DeepLIFT uses a zero baseline with $\epsilon =  10^{-10}$, and Integrated Gradients employs a zero baseline with 50 integration steps and Gauss--Legendre quadrature.

For the time series experiments, we evaluate RISE, LIME, KernelSHAP, and T-SHAP. {\color{black} Table \ref{tab:parameters_ts} depicted each method hyperparameters used in this case.}
RISE is configured with 1000 random masks generated from a coarse binary grid of resolution 16, with each cell activated with probability 0.5, followed by linear upsampling and random shifting to match the original sequence length. 
This configuration produces smooth perturbations while preserving local temporal structure. 
LIME is applied using the default tabular configuration, generating 5000 perturbed samples via Gaussian sampling without discretization. 
KernelSHAP also follows the default tabular configuration, with a background dataset constructed using the \textit{balanced centroid} strategy to ensure comparability with T-SHAP. 
T-SHAP is evaluated in both ROI and Window variants, with parameters set according to the original work~\cite{nguyen2025tshap}. 
In particular, the window length is set to 10\% of the sequence length, the stride to 5, and the background dataset is constructed using balanced class centroids.

For the tabular experiments, we evaluate LORE, LIME, Gradient, DeepLIFT, KernelSHAP, and Integrated Gradients. 
LIME and KernelSHAP use the same configuration adopted for binary images, with an occlusion value of 0, 1500 perturbed samples for LIME, and 25 samples for KernelSHAP. {\color{black} Table \ref{tab:parameters_tab} depicted each method hyperparameters used for this data typology.}
For gradient-based methods, Gradient is computed in absolute value, DeepLIFT uses a zero baseline with $\epsilon = 10^{-10}$, and Integrated Gradients employs a zero baseline with 50 integration steps and Gauss--Legendre quadrature. 
LORE~\cite{guidotti2019factual} is adapted for feature attribution using an impurity-based approach following~\cite{miro-nicolau2025comprehensive}, with neighborhood size set to 1000, 10 generations, and other-class ratio equal to 0.1.

\section{Results}
\label{sec:results}
We present here a comparative analysis of XAI methods using the SAIG frameworks introduced in Section~\ref{sec:method}, under the experimental setup described in the previous section. 
The goal is to assess the reliability of explanation methods across different data modalities. 
We begin with binary images, and subsequently extend the analysis to time series and tabular data, highlighting the strengths and limitations of each method in terms of explanation fidelity.

\begin{table}[t]
    \centering
    \begin{tabular}{lccc} 
    \toprule
                                                            & $\minf \uparrow$      & $\auc \uparrow$       & $\kl \downarrow$              \\ 
                                                            
                                                            \midrule
    RISE~\cite{petsiuk2018rise}                             & $0.343 \pm 0.382$     & $0.573 \pm 0.273$     & $1.915 \pm 1.505$             \\
    LIME~\cite{ribeiro2016why}                              & $0.293 \pm 0.164$     & $\mathbf{0.932 \pm 0.083}$     & $\mathbf{1.217 \pm 0.664}$             \\
    Gradient~\cite{simonyan2014deep}                        & $0.378 \pm 0.427$     & $0.500 \pm 0.026$     & $1.903 \pm 1.517$             \\
    DeepLIFT~\cite{shrikumar2017learning}                   & $\mathbf{0.380 \pm 0.429}$     & $0.555 \pm 0.123$     & $1.895 \pm 1.510$             \\
    KernelSHAP~\cite{lundberg2017unified}                   & $0.342 \pm 1.210$     & $0.848 \pm 0.165$     & $1.551 \pm 1.210$             \\
    Integrated Gradients~\cite{sundararajan2017axiomatic}   & $\mathbf{0.380 \pm 0.430}$     & $0.510 \pm 0.098$     & $1.905 \pm 1.519$             \\ 
    \bottomrule
    \end{tabular}
    \caption{Metrics results to compare XAI methods output and proposed Ground Truth in the Binary images task. All reported value are the mean and standard deviation for the test set. Best values for each metric in bold.}
    \label{tab:bin_img}
\end{table}

\begin{figure}[t]
    \centering
    \begin{subfigure}[b]{0.24\linewidth}
        \centering
        \includegraphics[width=\linewidth]{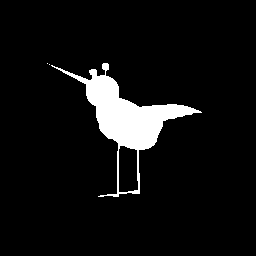}
        \caption{Input Image}
    \end{subfigure}
    \hfill
    \begin{subfigure}[b]{0.24\linewidth}
        \centering
        \includegraphics[width=\linewidth]{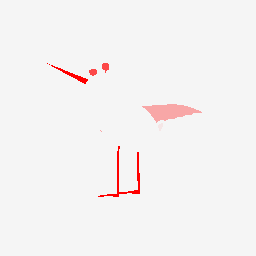}
        \caption{Pixel importance GT }
    \end{subfigure}
    \hfill
    \begin{subfigure}[b]{0.24\linewidth}
        \centering
        \includegraphics[width=\linewidth]{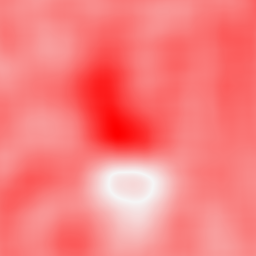}
        \caption{RISE}
    \end{subfigure}
    \hfill 
    \begin{subfigure}[b]{0.24\linewidth}
        \centering
        \includegraphics[width=\linewidth]{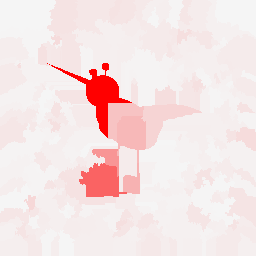}
        \caption{LIME}
    \end{subfigure}
    
    \begin{subfigure}[b]{0.24\linewidth}
        \centering
        \includegraphics[width=\linewidth]{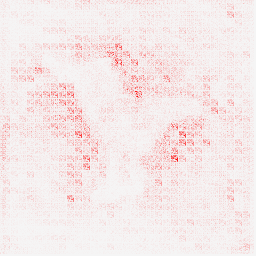}
        \caption{Grad-CAM}
    \end{subfigure}
    \hfill
    \begin{subfigure}[b]{0.24\linewidth}
        \centering
        \includegraphics[width=\linewidth]{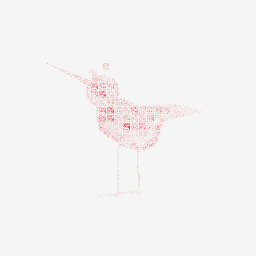}
        \caption{DeepLIFT}
    \end{subfigure}
    \hfill
    \begin{subfigure}[b]{0.24\linewidth}
        \centering
        \includegraphics[width=\linewidth]{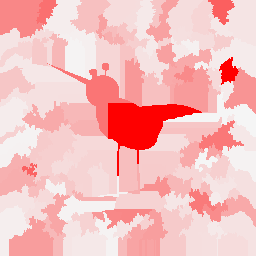}
        \caption{KernelSHAP}
    \end{subfigure}
    \hfill
    \begin{subfigure}[b]{0.24\linewidth}
        \centering
        \includegraphics[width=\linewidth]{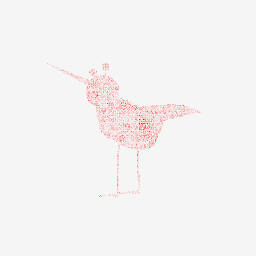}
        \caption{Int. Gradients}
    \end{subfigure}

    \caption{Comparison of different XAI saliency map techniques for the binary image dataset The more intense the color the more important the pixel.}
    \label{fig:bin_img}
\end{figure}

\paragraph{Binary Images}
Table~\ref{tab:bin_img} reports the comparison between the considered XAI methods and the corresponding ground truth explanations for binary images, using the evaluation metrics introduced earlier. 
The reported values correspond to the mean and standard deviation computed over the validation set for each method and metric. 
Figure~\ref{fig:bin_img} provides qualitative examples of the generated explanations.
As previously discussed, the $\minf$ metric primarily evaluates the spatial agreement between saliency maps. 
According to this measure, DeepLIFT~\cite{shrikumar2017learning}, Integrated Gradients~\cite{sundararajan2017axiomatic}, and Gradient~\cite{simonyan2014deep} emerge as the top-performing methods, yielding nearly identical results. 
This behavior is consistent with their shared underlying mechanism, as all three methods rely on backpropagating the model output to the input space. 
Nevertheless, their performance remains significantly below the theoretical optimum, i.e., $\minf = 1.0$, and shows only a marginal improvement over the lowest-performing method, LIME~\cite{ribeiro2016why}.
In contrast, the $\auc$ metric reveals a different behavior. 
All methods, except LIME~\cite{ribeiro2016why} and KernelSHAP~\cite{lundberg2017unified}, obtain values close to that of a random saliency map, i.e., $\auc = 0.5$. 
This can be explained by the fact that $\auc$ penalizes center-biased saliency maps~\cite{riche2013saliency}, a common characteristic of gradient-based methods. 
LIME and KernelSHAP, on the other hand, rely on the Quickshift segmentation algorithm~\cite{vedaldi2008quick} to isolate object regions from the background. As a result, they assign little or no importance to background pixels, which leads to higher $\auc$ scores. However, as indicated by the $\minf$ metric, these methods still fail to correctly attribute importance to specific object components.

Overall, the evaluated methods struggle to consistently align with the ground truth, underscoring the limitations of current XAI techniques even in relatively structured visual settings.

\begin{figure}[t]
    \centering
    \includegraphics[width=1\linewidth]{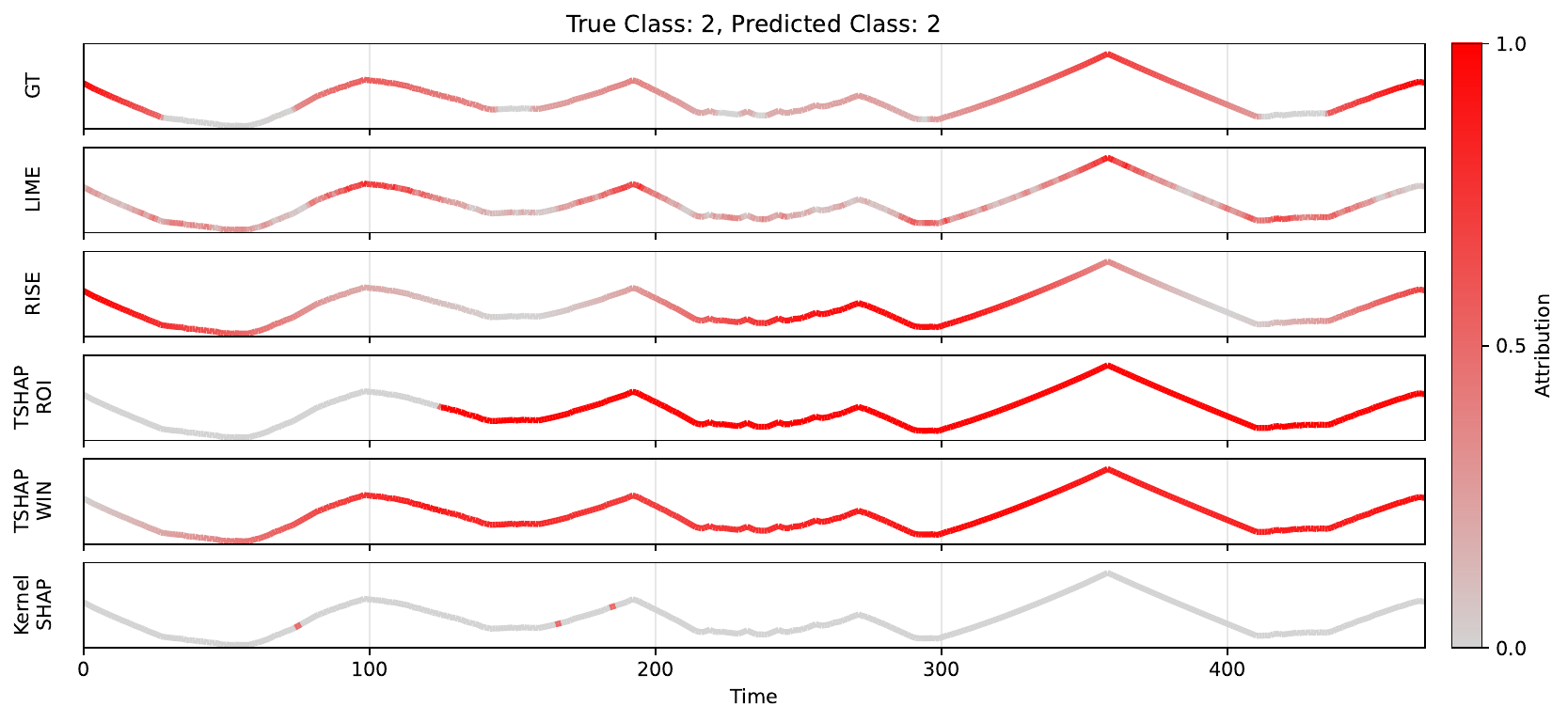}
    \caption{Time series attribution for an instance of class \textit{2}. From top to bottom, the ground truth (GT), and the predicted attributions from LIME, RISE, TSHAP (ROI and WIN), and KernelSHAP. The more intense the color the more important the time series observation.}
    \label{fig:attributionts}
\end{figure}

\paragraph{Time Series}
Figure~\ref{fig:attributionts} shows a representative example of time series attributions. 
From top to bottom, the figure reports the ground truth (GT) and the explanations produced by LIME, RISE, T-SHAP (ROI and Window), and KernelSHAP. 
Higher color intensity indicates greater importance. 
For comparability, all attribution maps are transformed by taking the absolute value and applying min--max normalization.

\begin{table}[t]
    \centering
    \begin{tabular}{lccc} \toprule
                            & $\minf \uparrow$      & $\auc \uparrow$       & $\kl \downarrow$              \\ \midrule
     RISE                    &  $\mathbf{0.522 \pm 0.136}$     & $\mathbf{0.709 \pm 0.161}$     & $\mathbf{0.790 \pm 0.405}$             \\
    LIME                    & $0.442 \pm 0.199$     & $0.646 \pm 0.134$     & $0.907 \pm 0.510$             \\
    Kernel SHAP             & $0.007 \pm 0.003$     & $0.500 \pm 0.003$     & $5.315 \pm 0.509$             \\
    T-SHAP ROI              & $0.386 \pm 0.170$     & $0.490 \pm 0.089$     & $1.784 \pm 0.767$             \\
    T-SHAP Window           & $0.424 \pm 0.191$     & $0.490 \pm 0.132$     & $1.055 \pm 0.562$             \\ \bottomrule
    \end{tabular}
    \caption{Metrics results to compare XAI methods output and proposed Ground Truth in the Time series task. All reported value are the mean and standard deviation for the validation set. Best values for each metric in bold.}
    \label{tab:ts}
\end{table}

Table~\ref{tab:ts} presents the comparison between XAI methods and ground truth explanations for the time series task, reporting mean and standard deviation values over the validation set for each metric.
According to the $\minf$ metric, RISE~\cite{petsiuk2018rise} achieves the best overall performance. 
This suggests that perturbation-based methods are particularly well suited for sequential data, as they directly estimate the effect of masking temporal segments on the model prediction. 
LIME~\cite{ribeiro2016why} also achieves competitive performance, indicating that local surrogate models can partially capture relevant temporal patterns.
KernelSHAP~\cite{lundberg2017unified} performs substantially worse than all other methods. 
The extremely low $\minf$ value indicates that the resulting importance distribution is almost uncorrelated with the ground truth, which is further confirmed by the high $\kl$ divergence. 
A possible explanation is that KernelSHAP treats each time step as an independent feature, failing to capture the strong temporal dependencies inherent to time series data. 
Also, the performance may be influenced by the choice of background dataset. 
In order to maintain computational feasibility, we adopt the balanced centroid strategy instead of a larger background set, which may reduce the representativeness of the reference distribution and negatively affect the explanations.
The two variants of T-SHAP exhibit intermediate performance. 
Both the ROI and Window strategies achieve moderate $\minf$ values, indicating that they are able to partially identify relevant temporal regions. 
However, their $\auc$ values remain close to $0.5$, suggesting a limited ability to correctly rank important and non-important time steps. 
Among the two variants, the Window strategy shows slightly better alignment with the ground truth distribution, as reflected by the lower $\kl$ divergence.
The $\auc$ metric further confirms that RISE and LIME outperform the remaining methods, indicating that perturbation-based and local surrogate approaches are more effective in isolating relevant temporal segments. Nevertheless, none of the evaluated methods closely matches the ground truth distribution, as evidenced by the relatively high $\kl$ values across all methods. Since $\kl$ evaluates the full probability distribution of the saliency map, it captures discrepancies in both central and peripheral regions, highlighting the difficulty of accurately reproducing the complete ground truth distribution.

\begin{table}[t]
    \centering
    \begin{tabular}{lcc} \toprule
                                                            & $\minf \uparrow$      & $\kl \downarrow$          \\ \midrule
    LORE~\cite{guidotti2019factual}                         & $0.784 \pm 0.177$     & $1.204 \pm 1.855$         \\ 
    LIME~\cite{ribeiro2016why}                              & $0.560 \pm 0.269$     & $3.806 \pm 3.535$         \\
    Gradient~\cite{simonyan2014deep}                        & $\mathbf{0.980 \pm 0.099}$     & $\mathbf{0.092 \pm 0.618}$         \\
    DeepLIFT~\cite{shrikumar2017learning}                   & $0.636 \pm 0.221$     & $2.011 \pm 2.559$         \\
    Kernel SHAP~\cite{lundberg2017unified}                  & $0.601 \pm 0.229$     & $2.315 \pm 2.701$         \\
    Integrated Gradients~\cite{sundararajan2017axiomatic}   & $0.642 \pm 0.221$     & $1.983 \pm 2.559$         \\ \bottomrule
    \end{tabular}
    \caption{Metrics results to compare XAI methods output and proposed Ground Truth in the Tabular data task. All reported value are the mean and standard deviation for the validation set. Best values for each metric in bold.}
    \label{tab:tabular}
\end{table}

\paragraph{Tabular Data}
Table~\ref{tab:tabular} reports the evaluation results for the tabular experiments. 
As discussed in Section~\ref{sec:experimental}, the $\auc$ metric is not considered in this setting due to the absence of spatial structure.
Compared to the binary image and time series tasks, we observe a general improvement in the $\minf$ metric. 
This is likely due to the lower complexity of the tabular setting. LIME~\cite{ribeiro2016why} obtains the worst performance across both evaluation metrics. 
In contrast, Gradient~\cite{simonyan2014deep} achieve near-perfect $\minf$ scores. The second-best method, according to $\minf$ measure, is LORE~\cite{guidotti2019factual}. Integrated 
The $\kl$ metric confirms this trend, with the same methods obtaining the best results. 
However, KernelSHAP performs slightly worse than the other two methods, which achieve values very close to the optimum.

\section{Conclusion}
\label{sec:conclusions}
In this work, we have introduced a unified intervention-based framework for generating Synthetic Artificial Intelligence Ground Truth (SAIG) across multiple data modalities, namely binary images, tabular data, and time series. 
In particular, we presented, to the best of our knowledge, the first local ground truth SAIG construction for time series data tabular data. 
By extending intervention-based principles beyond the image domain, the proposed framework enables the systematic and controlled evaluation of XAI methods under settings where the true contribution of each input component is known by design.
The experimental results highlight several limitations of current XAI techniques. 
Across all modalities, none of the evaluated methods consistently achieves high fidelity with respect to the ground truth explanations. 
While gradient-based methods perform relatively well in structured settings such as images and tabular data, their performance degrades when evaluated using stricter distribution-based metrics. 
Perturbation-based approaches, such as RISE, show stronger performance in time series data, suggesting that different methodological families are better suited to different data characteristics. 
These findings reinforce the idea that explanation quality is highly dependent on the underlying data modality and evaluation setting. 
In this context, our results provide empirical support for the \textit{No Free Lunch Theorem for explanations}~\cite{han2022which}, indicating that no single XAI method can be expected to perform optimally across all tasks.

Despite these contributions, the proposed approach presents some limitations. 
As with existing SAIG methodologies, the evaluation is conducted in synthetic environments, where the data generation process is controlled. 
While this design enables the definition of ground truth explanations, it may not fully capture the complexity and variability of real-world scenarios. 
Hence, the insights obtained from such benchmarks should be interpreted as complementary to, rather than a replacement for, evaluations performed in real applications.
Future research directions include extending the proposed framework to more complex and realistic data generation processes, as well as exploring hybrid evaluation strategies that combine synthetic ground truth with real-world validation. 
Furthermore, the development of standardized SAIG benchmarks could facilitate more consistent and reproducible comparisons across XAI methods, ultimately contributing to the design of more robust and reliable explanation techniques.

\printcredits

\bibliographystyle{cas-model2-names}


\bibliography{refs}


\bio{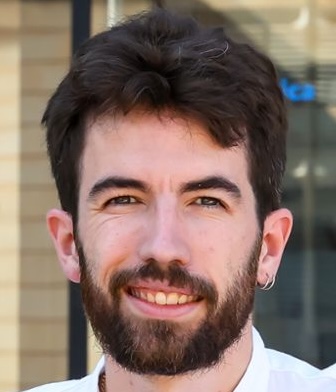}
Assistant lecturer with a PhD, member of LAIA@UIB, UGIVIA, and SCOPIA research groups. from Universitat de les Illes Baleares.  His current research builds on the foundations laid during his PhD, with the aim of further improving the evaluation of post-hoc XAI techniques. He seeks to enrich the field by integrating insights from related areas such as social sciences, adversarial robustness, and the challenges posed by out-of-distribution data
\endbio

\vskip3pc

\bio{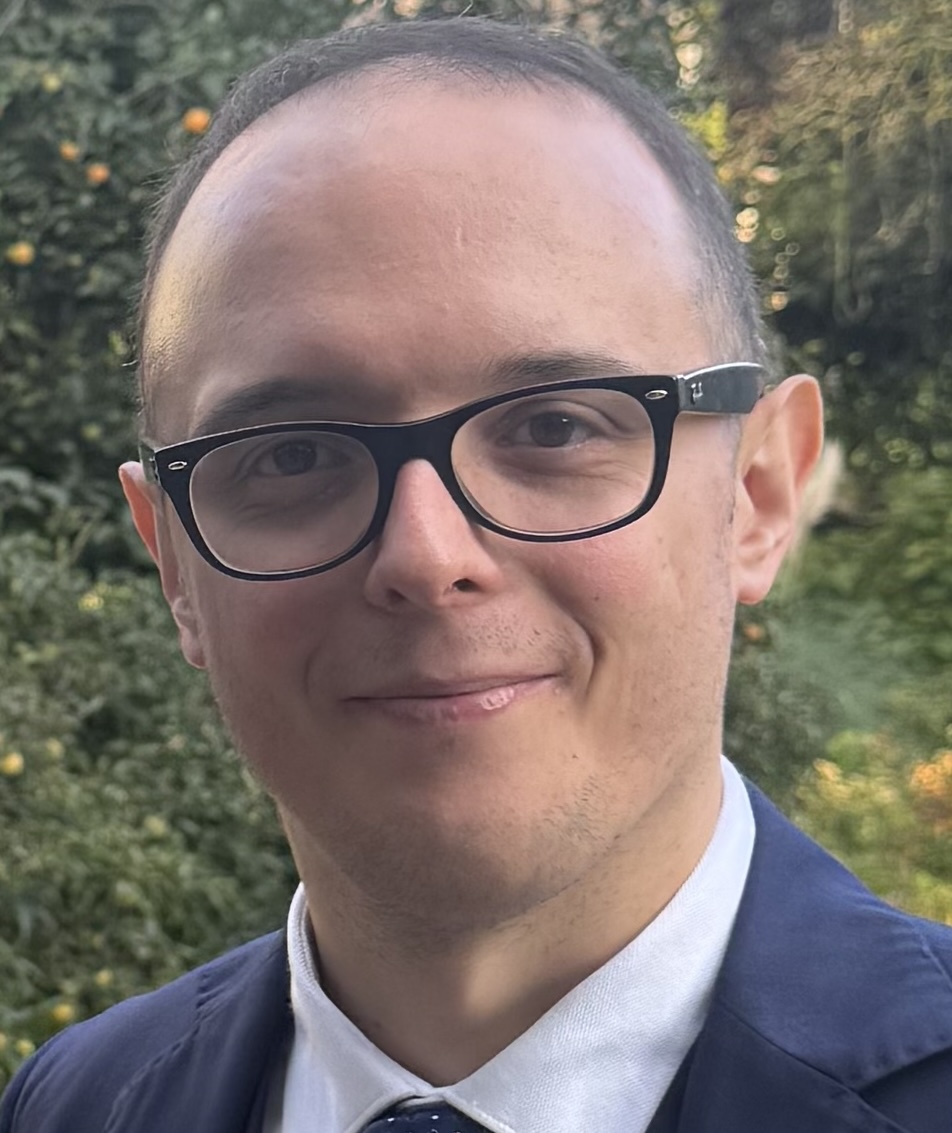}
Francesco Spinnato is a researcher and junior lecturer at the University of Pisa working on explainable artificial intelligence for sequential data, with a focus on interpretable methods for time series and black-box models. He received his Ph.D. in Data Science from the Scuola Normale Superiore in 2024. His research interests include XAI, interpretable machine learning, time series analysis, and sequential data modeling, with applications in domains such as mobility, health, and insurance.
\endbio

\vskip3pc

\bio{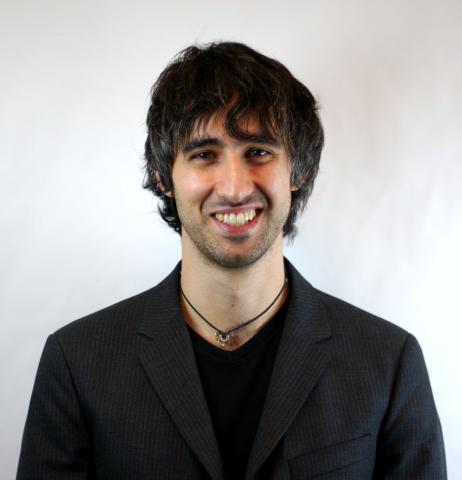}
Riccardo Guidotti is an Associate Professor at University of Pisa. In 2013 and 2010 he graduated cum laude in Computer Science (MS and BS) at University of Pisa.
He received the PhD in Computer Science with a thesis on Personal Data Analytics in the same institution. He is currently an Associate Professor at the Department of Computer Science University of Pisa, Italy, and a member of the Knowledge Discovery and Data Mining Laboratory (KDDLab), a joint research group with the Information Science and Technology Institute of the National Research Council in Pisa. He won the IBM fellowship program and has been an intern in IBM Research Dublin, Ireland in 2015. He also won the DSAA New Generation Data Scientist Award 2018, and the Marco Somalvico Award 2021. His research interests are in explainable artificial intelligence, interpretable machine learning, quantum computing, fairness, and bias detection, time series analysis, data generation, personal data mining, clustering, and analysis of transactional data.
\endbio

\end{document}